%% file: main.tex
\documentclass[10pt,twocolumn,letterpaper]{article}

\usepackage{cvpr}
\usepackage{times}
\usepackage{epsfig}
\usepackage{graphicx}
\usepackage{amsmath}
\usepackage{amssymb}
\usepackage{soul}

\usepackage{booktabs}
\usepackage{xspace}
\usepackage{subcaption}
\usepackage{verbatim}

\usepackage[pagebackref=true,breaklinks=true,letterpaper=true,colorlinks,bookmarks=false]{hyperref}

\cvprfinalcopy %

\def\cvprPaperID{6354} %

\ifcvprfinal\fi
\begin{document}

\title{Interactive Full Image Segmentation by Considering All Regions Jointly}

\author{Eirikur Agustsson\\
Google Research\\
{\tt\small eirikur@google.com}
\and
Jasper R. R. Uijlings\\
Google Research\\
{\tt\small jrru@google.com}
\and
Vittorio Ferrari\\
Google Research\\
{\tt\small vittoferrari@google.com}
}

\maketitle
\thispagestyle{empty}

\input{macros}

\input{abstract}
\input{introduction}

\input{related_work}

\input{method}

\input{results}

\input{discussion}
\input{conclusion}

{\small
\bibliographystyle{ieee}
\bibliography{shortstrings,loco}
}

\end{document}

%% file: macros.tex
\newcommand{\cls}[1]{\texttt{#1}}  %
\newcommand{\fbseg}{foreground/background segmentation\xspace}
\newcommand{\stuff}{stuff\xspace}  %
\newcommand{\full}{\texttt{full image}\xspace}
\newcommand{\single}{\texttt{single region}\xspace}
\newcommand{\singlescrib}{\texttt{single region scribble}\xspace}
\newcommand{\fullscrib}{\texttt{full image scribble}\xspace}
\newcommand{\fullone}{\texttt{full image, one scribble per region}\xspace}

\newif\ifdraft
\drafttrue
\ifdraft
  \newcommand{\eirikur}[1]{{\color{blue}{#1}}}
  \newcommand{\jasper}[1]{{\color{magenta}{#1}}}  
  \newcommand{\vitto}[1]{{\color{red}{V: #1}}}
\else
  \newcommand{\eirikur}[1]{}
  \newcommand{\jasper}[1]{}
  \newcommand{\vitto}[1]{}
\fi

\newcommand{\mypartop}[1]{\vspace{0mm}{\noindent\textbf{#1}}}
\newcommand{\mypar}[1]{\vspace{0mm}{\noindent\textbf{#1}}}

\newcommand{\image}{\mathbf{X}}
\newcommand{\pixel}{{X}}
\newcommand{\regionlabels}{\mathbf{Y}} %
\newcommand{\regionlabel}{{Y}} %
\newcommand{\singleregionlabel}{{y}} %
\newcommand{\classlabels}{\mathbf{c}} %
\newcommand{\classlabel}{{c}} %
\newcommand{\features}{\mathbf{Z}}  %
\newcommand{\cfeatures}{\mathbf{S}}  %
\newcommand{\rfeatures}{\mathbf{z}} %
\newcommand{\rcfeatures}{\mathbf{s}}%
\newcommand{\postroifeatures}{\mathbf{v}}%
\newcommand{\rlogits}{\mathbf{l}}  %
\newcommand{\logits}{\mathbf{L}}   %
\newcommand{\logit}{{L}}   %
\newcommand{\probs}{\mathbf{P}}   %
\newcommand{\prob}{{P}}   %
\newcommand{\boxes}{\mathbf{b}} %
\newcommand{\boxe}{\mathbf{b}} %
\newcommand{\softmax}[1]{\text{softmax}(#1)}
\newcommand{\xy}{{(x,y)}} %
\newcommand{\pixelloss}{\mathcal{L}_{\text{pixelwise}}}

%% file: abstract.tex
\begin{abstract}

We address interactive full image annotation, where the goal is to accurately segment all object and stuff regions in an image.
We propose an interactive, scribble-based annotation framework which operates on the whole image to produce segmentations for all regions. This enables sharing scribble corrections across regions, and allows the annotator to focus on the largest errors made by the machine across the whole image.
To realize this, we adapt Mask-RCNN~\cite{he17iccv} into a fast interactive segmentation framework and introduce an instance-aware loss measured at the pixel-level in the full image canvas, which lets predictions for nearby regions properly compete for space.
Finally, we compare to interactive single object segmentation on the COCO panoptic dataset~\cite{caesar18cvpr,kirillov18arxiv,lin14eccv}. We demonstrate that our interactive full image segmentation approach leads to a 5\% IoU gain, reaching 90\% IoU at a budget of four extreme clicks and four corrective scribbles per region.
\end{abstract}

%% file: introduction.tex
\section{Introduction}

We address the task of interactive full image segmentation, where the goal is to obtain accurate segmentations for all object and \stuff regions in the image. 
Full image annotations are important for many applications such as self-driving cars~\cite{cordts16cvpr,geiger13ijrr}, navigation assistance for the blind~\cite{serrao15infosociety}, and automatic image captioning~\cite{johnson16cvpr,wang17cvpr}.
However, creating such datasets requires large amounts of human labor. For example, annotating a single image took 1.5 hours for Cityscapes~\cite{cordts16cvpr}. For COCO+stuff~\cite{caesar18cvpr,lin14eccv}, annotating one image took 19 minutes (80 seconds per object~\cite{lin14eccv} plus 3 minutes for \stuff regions~\cite{caesar18cvpr}), which totals 39k hours for the 123k images. So there is a clear need for faster annotation tools.

This paper proposes an efficient interactive framework for full image segmentation (Fig.~\ref{fig:main_idea} and~\ref{fig:architecture}). Given an image, an annotator first marks extreme points~\cite{papadopoulos17iccv} on all object and \stuff regions. These provide a tight bounding box with four boundary points for each region, and can be efficiently collected (7s per region~\cite{papadopoulos17iccv}).
Next, the machine predicts an initial segmentation for the full image based on these extreme points.
Afterwards we present the whole image with the predicted segmentation to the annotator and iterate between
(A) the annotator providing scribbles on the errors of the current segmentation, and
(B) the machine updating the predicted segmentation accordingly (Fig.~\ref{fig:main_idea}).

Our approach of full image segmentation brings several advantages over modern interactive single object segmentation methods
~\cite{hu19nn,le18eccv,li18cvpr,liew17iccv,mahadevan18bmvc,maninis18cvpr,xu16cvpr}:
(I) It enables the annotator to focus on the largest errors in the whole image, rather than on the largest error of one given object.
(II) It shares annotations across multiple object and stuff regions. In our approach, a single scribble correction specifies the extension of one region and the shrinkage of neighboring regions (Sec.~\ref{sec:annotation_features} and Fig.~\ref{fig:conditioning}). In interactive single object segmentation, corrections are used for the given target object only.
(III) Our approach lets regions compete for space in the common image canvas, ensuring that a pixel is assigned exactly one label (Fig.~\ref{sec:architecture}). In single object segmentation instead, pixels along boundary regions may be assigned to multiple objects, leading to contradictory labels, or to none, leading to holes. At the same time, since regions compete, corrections of one region influences nearby regions in our framework (e.g.~Fig.~\ref{fig:qualitative_examples}).
(IV) Instead of only annotating object instances, we also annotate \stuff regions, capturing important classes such as \cls{pavement} or \cls{river}.

We realize interactive interactive full image segmentation by adapting Mask-RCNN~\cite{he17iccv} (Fig.~\ref{fig:architecture}). We start from extreme points~\cite{papadopoulos17iccv}, which define bounding boxes. Therefore we bypass the Region Proposal Network of Mask-RCNN and use these boxes directly to extract Region-of-Interest (RoI) features (Sec.~\ref{sec:architecture}).
Afterwards, we incorporate extreme points and scribble annotations inside Mask-RCNN by concatenating them to the RoI-features. We encode annotations in a way that allows to share them across regions, enabling advantage (II) above (Sec.~\ref{sec:annotation_features}).
Finally, while Mask-RCNN~\cite{he17iccv} predicts each mask separately, we project the mask predictions back on the pixels in the common image canvas~\ref{sec:architecture}. Then we define a new loss which is instance-aware yet lets predictions properly compete for space, enabling advantage (III) above (Sec.~\ref{sec:loss}).

To the best of our knowledge, all deep interactive single object segmentation methods~\cite{hu19nn,le18eccv,li18cvpr,liew17iccv,mahadevan18bmvc,maninis18cvpr,xu16cvpr} are based on Fully Convolutional Networks (FCNs)~\cite{chen18pami,long15cvpr,ronneberger15miccai}. We chose to start from Mask-RCNN~\cite{he17iccv} for efficiency. FCN-style interactive segmentation methods concatenate corrections to a crop of an RGB image, and pass that through a large neural net (e.g. ResNet-101~\cite{he16cvpr}). This requires a full inference pass for each region at each correction iteration.
In our Mask-RCNN framework instead, the RGB image is first passed through the large backbone network. Afterwards, for each region only a pass over the final segmentation head is required (Fig.~\ref{fig:architecture}). This is much faster and more memory efficient (Sec.~\ref{sec:discussion}).

We perform thorough experiments in increasingly complex settings:
(1) {\em Single object segmentation}: On the COCO dataset~\cite{lin14eccv}, our Mask-RCNN style architecture achieves similar performance to DEXTR~\cite{maninis18cvpr} on single object segmentation from extreme points~\cite{papadopoulos17iccv}.
(2) {\em Full image segmentation}: We evaluate on the COCO panoptic challenge~\cite{caesar18cvpr,kirillov18arxiv,lin14eccv} the task of segmenting all object and stuff regions in an image, starting from extreme points.
Our idea to share annotations across regions in combination with our pixel-wise loss yield a +3\% IoU gain over an interactive single region segmentation baseline.
(3) {\em Interactive full image segmentation}: On the COCO panoptic challenge, we demonstrate the combined effects of our three advantages (I)-(III) above: at a budget of four extreme clicks and four scribbles per region, we get a total +5\% IoU gain over the interactive single region segmentation baseline.

%% file: related_work.tex
\begin{figure*}[t]
	\vspace{-1.0cm}
    \hspace{-.7cm}
	\includegraphics[width=1.1\linewidth]{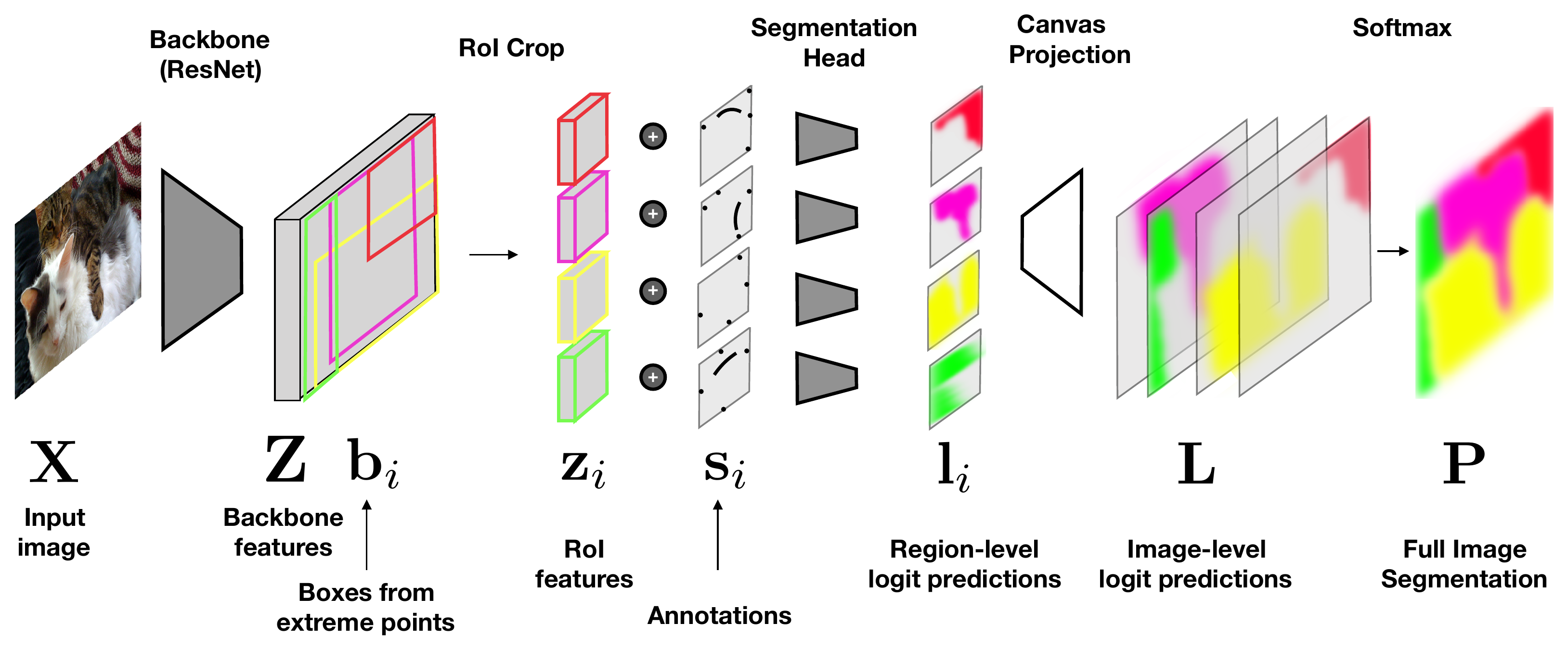}
	\vspace{-0.7cm}
	\caption{\small{Our proposed region based model for interactive full image segmentation (see Sec.~\ref{sec:architecture} for details).
We start from Mask-RCNN~\cite{he17iccv}, but use user provided boxes (from extreme points) instead of a box proposal networks for RoI cropping, and concatenate the RoI features with annotator provided corrective scribbles.
Instead of predicting binary masks for each region, we project all region prediction into the common image canvas, where they compete for space.
The network is trained end-to-end for a novel pixel-wise loss for the full image segmentation task (see Sec.~\ref{sec:loss})}.
}
    \vspace{0cm}
	\label{fig:architecture}
\end{figure*}
\section{Related Work}

\mypartop{Semantic segmentation from weakly labeled data.}
Many works address semantic segmentation by training from weakly labeled data, such as image-level labels~\cite{kolesnikov16eccv,pathak15iccv,wei18cvpr},
point-clicks~\cite{bearman16eccv,bell15cvpr,chen18aaai,wang14cviu}, boxes~\cite{khoreva17cvpr,maninis18cvpr,papadopoulos17iccv} and scribbles~\cite{lin16cvpr,xu15cvpr}. 
Boxes can be efficiently annotated using extreme points~\cite{papadopoulos17iccv} which can also be used as an extra signal for generating segmentations~\cite{maninis18cvpr,papadopoulos17iccv}.
This is related as our method starts from extreme points for each region.
However, the above methods operate from annotations collected before any machine processing. Our work instead is in the interactive scenario, where the annotator iteratively provide corrective annotations for the current machine segmentation.

\mypar{Interactive object segmentation.}
Interactive object segmentation is a long standing research topic. Most classical approaches~\cite{bai09ijcv,batra11ijcv,boykov01iccv,rother04siggraph,criminisi10tog,cheng15cgf,gulshan10cvpr,nagaraja15iccv} formulate object segmentation as energy minimization on a regular graph defined over pixels, with unary potential capturing low-level appearance properties and pairwise or higher-order potentials encouraging regular segmentation outputs.

Starting from Xu et al.~\cite{xu16cvpr}, recent methods address interactive object segmentation with deep neural networks~\cite{hu19nn,le18eccv,li18cvpr,liew17iccv,mahadevan18bmvc,maninis18cvpr,xu16cvpr}. These works build on Fully Convolutional architectures such as FCNs~\cite{long15cvpr} or Deeplab~\cite{chen18pami}. They input the RGB image plus two extra channels for object and non-object corrections, and output a binary mask.

In~\cite{chen18cvpr} they perform interactive object segmentation in video. They use Deeplab~\cite{chen18pami} to create a pixel-wise embedding space. Annotator corrections are used to create a nearest neighbor classifier on top of this embedding, enabling quick updates of the object predictions.

Finally, Polygon-RNN~\cite{acuna18cvpr,castrejon17cvpr} is an interesting alternative approach. Instead of predicting a mask, they uses a recurrent neural net to predict polygon vertices.
Corrections made by the annotator are used by the machine to refine its vertex predictions. 

\mypar{Interactive full image segmentation.}
Recently,~\cite{andriluka18acmmm} proposed Fluid Annotation, which also addresses the task of full image annotation. Our work shares the spirit of focusing annotator effort on the biggest errors made by the machine across the whole image. However,~\cite{andriluka18acmmm} uses Mask-RCNN~\cite{he17iccv} to create a large pool of fixed segments and then provides an efficient interface for the annotator to rapidly {\em select which of these} should form the final segmentation. In contrast, in our work all segments are created from the initial extreme points and are all part of the final annotation. Our method then enables to {\em correct the shape} of segments to precisely match object boundaries.

Several older works on interactive segmentation handle multiple labels in a single image~\cite{nieuwenhuis13pami,nieuwenhuis14eccv,santner10accv,vezhnevets05graphicon}. We present the first interactive deep learning framework which does this. Moreover, in contrast to those works, we explicitly demonstrate the benefits of interactive full image segmentation over interactive single object segmentation.

\mypar{Other works on interactive annotation.}
In~\cite{rupprecht18cvpr} they combine a segmentation network with a language module to allow a human to correct the segmentation by typing feedback in natural language, such as ``there are no clouds visible in this image''.
The work of~\cite{papadopoulos16cvpr} annotates bounding boxes using only human verification, while~\cite{konyushkova18cvpr} trained agents to determine whether it is more efficient to verify or draw a bounding box.
The avant-garde work of~\cite{russakovsky15cvpr} had a machine dispatching many labeling questions to annotators, including whether an object class is present, box verification, box drawing, and finding missing instances of a particular class in the image.
In~\cite{vijayanarasimhan09cvpr} they estimate the informativeness of having an image label, a box, or a segmentation for an image, which they use to guide an active learning scheme.
Finally, several works tackle fine-grained classification through attributes interactive provided by annotators~\cite{branson10eccv,parkash12eccv,biswas13cvpr,wah14cvpr}.

%% file: method.tex
\section{Our interactive segmentation model}\label{sec:model}

This section describes our model which we use to predict a segmentation from extreme points and scribble corrections (Fig.~\ref{fig:main_idea}).
We first discuss the model architecture (Sec.~\ref{sec:architecture}).
We then describe how we feed annotations to the model (extreme points and scribble corrections, Sec.~\ref{sec:annotation_features}).
Finally, we describe model training with our new loss function (Sec.~\ref{sec:loss}).

\vspace{0cm}
\subsection{Model architecture}\label{sec:architecture}

Our model is based on Mask-RCNN~\cite{he17iccv}. In Mask-RCNN inference is done as follows:
(1) An input image $\pixel$ is passed through a deep neural network backbone such as ResNet~\cite{he16cvpr}, producing a feature map $\features$.
(2) A specialized network module (RPN~\cite{ren15nips}) predicts box proposals based on $\features$. %
(3) These box proposals are used to crop out Region-of-Interest (RoI) features $\rfeatures$ from $\features$ with a RoI cropping layer (RoI-align~\cite{he17iccv}). 
(4) Then each RoI feature $\rfeatures$ is fed into three separate network modules which predict a class label, refined box coordinates, and a segmentation mask.

Fig.~\ref{fig:architecture} illustrates how we adapt Mask-RCNN~\cite{he17iccv} for interactive full image segmentation. In particular, our network takes three types of inputs:
(1) an image $\pixel$ of size $W \times H \times 3$;
(2) $N$ annotation maps $\cfeatures_1, \cdots, \cfeatures_N$ of size $W \times H$ (for extreme points and scribble corrections, Sec.~\ref{sec:annotation_features}); and
(3) $N$ boxes $\boxe_1, \cdots, \boxe_N$ determined by the extreme points provided by annotators. Here $N$ is the number of regions that we want to segment, which is determined by the annotator, and which may vary per image.

As in Mask-RCNN, an image $\pixel$ is fed into our backbone architecture (ResNet~\cite{he16cvpr}) to produce feature map Z of size $\frac{1}{r}W \times \frac{1}{r}H \times C$, where $C$ is the number of feature channels and $r$ is a reduction factor. Both $C$ and $r$ are determined by the choice of backbone architecture. %

In contrast to Mask-RCNN, we already have boxes $\boxe_1, \cdots, \boxe_N$, so we do not need a box proposal module. Instead, we use each box $\boxe_i$ directly to crop out an RoI feature $\rfeatures_i$ from feature map $\features$. All features $\rfeatures_i$ have the same fixed size $w \times h \times C$ (i.e. $w$ and $h$ are only dependent on the RoI cropping layer).
We concatenate to this the corresponding annotation map $\rcfeatures_i$ which is described in Sec.~\ref{sec:annotation_features}, and obtain a feature map $\postroifeatures_i$, which is of size $w \times h \times (C + 2)$.

Using $\postroifeatures_i$, our network predicts a logit map $\rlogits_i$ of size $w' \times h'$ which represents the prediction of a single mask.
While Mask-RCNN stops at such mask predictions and processes them with a sigmoid to obtain binary masks, we want to have predictions influence each other.
Therefore we use the boxes $\boxe_1, \cdots, \boxe_N$ to re-project the logit predictions of all masks $\rlogits_i$ back into the original image resolution which results in $N$ prediction maps $\logits_i$.
We concatenate these prediction maps into a single tensor $\logits$ of size $W \times H \times N$.
For each pixel, we then obtain region probabilities $\probs$ of dimension $W\times H \times N$ by applying a softmax to the logits,
\begin{equation}
	(\prob^\xy_1,\cdots,\prob^\xy_N) = \softmax{\logit^\xy_1,\cdots,\logit^\xy_N},
\end{equation}
where $\prob^\xy_i$ denotes the probability that pixel $\xy$ is assigned to region $i$.
This makes multiple nearby regions compete for space in the common image canvas.

\vspace{0cm}
\subsection{Incorporating annotations}\label{sec:annotation_features}

\begin{figure*}[thb]
\centering
\vspace{-0.6cm}
    \begin{minipage}{0.63\linewidth}
    \includegraphics[width=\linewidth]{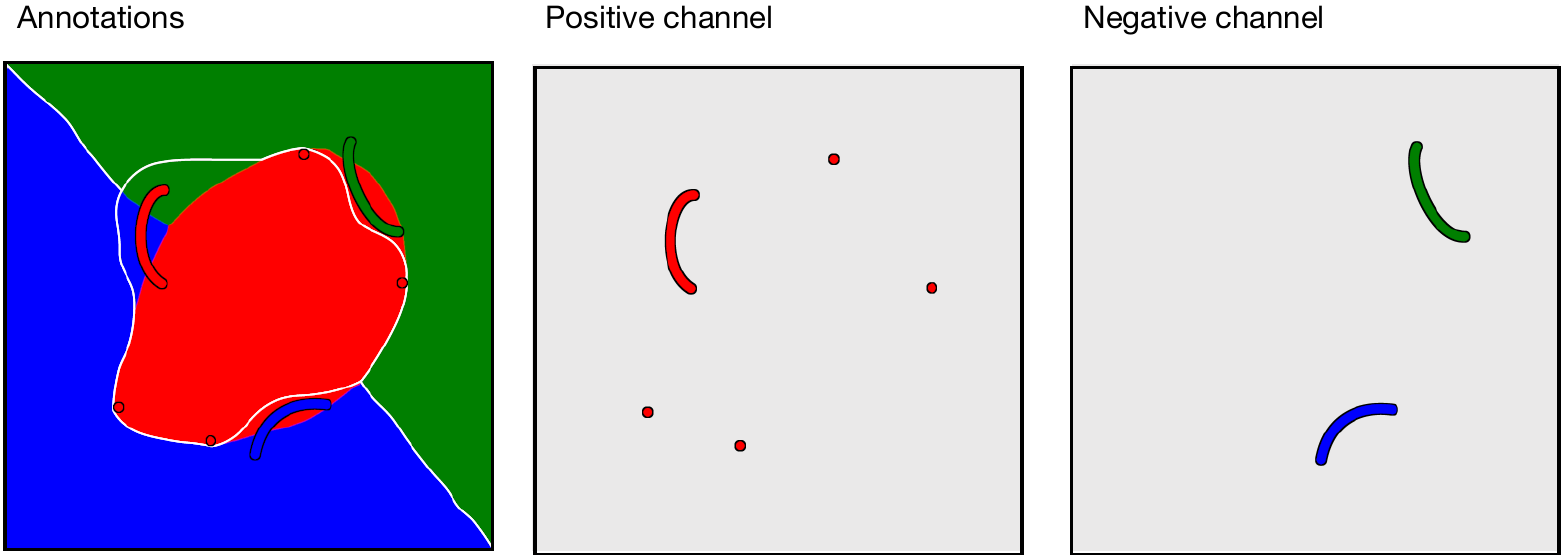}
    \caption{\label{fig:conditioning}\small{We illustrate how we combine all annotations near a region (red) into two annotation maps specific for that region. The colored regions denote the current predicted segmentation and the white boundaries depict true object boundaries. For the red region, the extreme points and the single positive scribble are combined into a single positive binary channel. All scribbles from other nearby regions are collected into a single negative binary channel. }}
\end{minipage}
\hspace{0.5cm}
\begin{minipage}{0.325\linewidth}
\centering
    \includegraphics[width=0.73\linewidth]{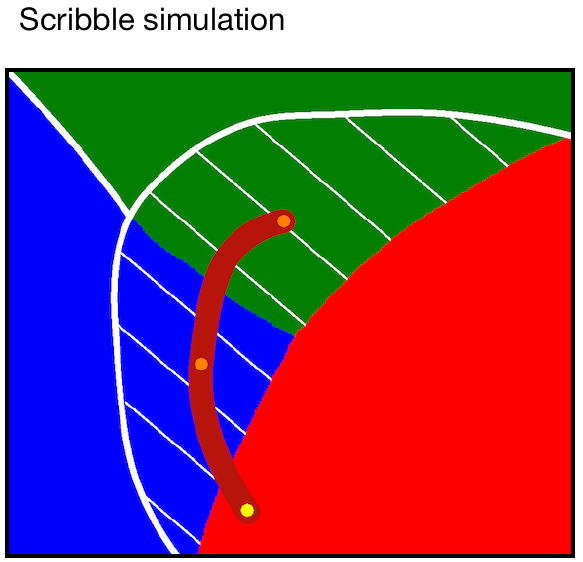}
    \caption{\label{fig:error_region}\small{
    To simulate a corrective scribble, we first sample an initial control point to indicate which region we want to expand (yellow), followed by two control points (orange) sampled uniformly from the error region.
    }}
\end{minipage}
\end{figure*}

Our model in Fig.~\ref{fig:architecture} concatenates RoI features $\rfeatures$ with annotation map $\rcfeatures$. We now describe how we create $\rcfeatures$. First, for each region $i$ we create a positive annotation map $\cfeatures_i$ which is of the same size $W \times H$ as the image. We choose the annotation map to be binary and we create it by pasting all extreme points and corrective scribbles for region $i$ onto it. Extreme points are represented by a circle which is 6 pixels in diameter. Scribbles are 3 pixels wide. 

For each region $i$, we collapse all annotations which do not belong to it into a single negative annotation map $\sum_{j\neq i} \cfeatures_j$. Then, we concatenate the positive and negative annotation maps into a two-channel annotation map $F_i$ 
\begin{equation}
F_i:=\Big(\cfeatures_i, \;\; \sum_{j\neq i} \cfeatures_j \Big),
\end{equation} which is illustrated in Fig.~\ref{fig:conditioning}.
Finally, we apply RoI-align~\cite{he17iccv} to $F_i$ using box $\boxe_i$ to obtain the desired cropped annotation map $\rcfeatures_i$.

The way we construct $F_i$ enables the sharing of all annotation information across multiple object and stuff regions in the image. The negative annotations for one region are formed by collecting the positive annotations of all other regions.
In contrast, in single object segmentation works~\cite{bai09ijcv,boykov01iccv,criminisi10tog,cheng15cgf,gulshan10cvpr,hu19nn,le18eccv,li18cvpr,liew17iccv,nagaraja15iccv,mahadevan18bmvc,maninis18cvpr,rother04siggraph,xu16cvpr} both positive and negative annotations are made only on the target object and they are never shared, so they only have an effect on that one object.

\subsection{Training}\label{sec:loss}

\mypartop{Training data.}
As training data, we have ground-truth masks for all objects and stuff regions in all images.
We represent the (non-overlapping) $N$ ground truth masks of an image $\image$ with region indices. This results in a map $\regionlabels$ of dimension $W\times H$, which assigns each pixel $\pixel^\xy$ to a region $\regionlabel^\xy\in \{1,...N\}$.

\mypar{Pixel-wise loss.}
Standard Mask-RCNN is trained with Binary Cross Entropy (BCE) losses for each mask prediction separately.
This means that there is no direct interaction between adjacent masks, and they might even overlap. Instead, we propose a novel instance-aware loss which lets predictions compete for space in the original image canvas.

In particular, as described in Sec.~\ref{sec:architecture} we project all region-specific logits into a single image-level logit tensor $\logits$, which is softmaxed into a region assignment probabilities $\probs$ of size $W\times H\times N$. 

As described above, the ground-truth segmentation is represented by $\regionlabels$ with values in $\{1, \cdots, N\}$, which specifies for each pixel its region index.  Since we simulate the extreme points from the ground-truth masks, there is a direct correspondence between the region assignment probabilities $\probs_1,\cdots,\probs_N$ and $\regionlabels$.
Thus, we can train our network end-to-end for the Categorical Cross Entropy (CCE) loss for the region assignments:
\begin{equation}
\pixelloss = \sum_{\xy} - \log P_{Y^\xy}^\xy \label{eq:pixelloss}
\end{equation}
We note that while the CCE loss is commonly used in fully convolutional networks for semantic segmentation~\cite{chen18pami,long15cvpr,ronneberger15miccai}, we instead use it in an architecture based on Mask-RCNN~\cite{he17iccv}.
Furthermore, usually the loss is defined over a fixed number of classes ~\cite{chen18pami,long15cvpr,ronneberger15miccai}, whereas we define it over the number of regions $N$. This number of regions may vary per image.

The loss in \eqref{eq:pixelloss} is computed over the pixels in the full resolution common image canvas.
Consequently, larger regions have a greater impact on the loss. However, in our experiments we measure Intersection-over-Union (IoU) between ground-truth masks and predictions, which considers all regions equally independent of their size. 
Therefore we weigh the terms in \eqref{eq:pixelloss} as follows. For each pixel we find the smallest box $\boxe_i$ which contains it, and reweigh the loss for that pixel by the inverse of the size of $\boxe_i$.
This causes each region to contribute to the loss approximately equally.

Our loss shares similarities with~\cite{caesar16eccv}. They used Fast-RCNN~\cite{girshick14cvpr} with selective search regions~\cite{uijlings13ijcv} and generate a class prediction vector for each region. Then they project this vector back into the image canvas using its corresponding region, while resolving conflicts using a max operator. In our work instead, we project a full logit map back into the image (Fig.~\ref{fig:architecture}). Furthermore, while in~\cite{caesar16eccv} the number of logit channels is equal to the number of classes $C$, in our work it depends on the number of regions $N$, which may vary per image.

\vspace{0cm}
\subsection{Implementation details}

The original implementation of Mask-RCNN~\cite{he17iccv} creates for each RoI feature mask predictions for all classes that it is trained on. At inference time, it uses the predicted class to select the corresponding predicted mask. Since we build on Mask-RCNN, we also do this in our framework for convenience of the implementation. During training we use the class labels to train class-specific mask prediction logits. During inference, for each region $i$ we use the class label predicted by Mask-RCNN to select which mask logits which we use as $\rlogits_i$. Hence during inference time, we have implicit class labels. However, class labels are never exposed to the annotator and are considered to be irrelevant for this paper.

\vspace{0cm}
\section{Annotations and their simulation}\label{sec:annotator}

Our annotations consists of both extreme points and scribble corrections. We chose scribble corrections~\cite{bai09ijcv,boykov01iccv,rother04siggraph} instead of click corrections~\cite{hu19nn,le18eccv,li18cvpr,liew17iccv,mahadevan18bmvc,maninis18cvpr,xu16cvpr} as they are a more natural choice in our scenario. As we consider multiple regions in an image, any annotation first needs to indicate which region should be extended. With scribbles one can start inside the region to be extended, followed by a path which specifies how to extend the region.

In all our experiments we simulate annotations, following previous interactive segmentation works~\cite{acuna18cvpr,castrejon17cvpr,hu19nn,le18eccv,li18cvpr,liew17iccv,mahadevan18bmvc,maninis18cvpr,xu16cvpr}.

\mypartop{Simulating extreme points.}
To simulate the extreme points that the annotator provides at the beginning, we use the code provided by~\cite{maninis18cvpr}.

\mypar{Simulating scribble corrections.}
To simulate scribble corrections during the interactive segmentation process, we first need to select an error region. Error regions are defined as a connected group of pixels of a ground-truth region which has been wrongly assigned to a different region (Fig.~\ref{fig:error_region}). We assess the importance of an error region by measuring how much segmentation quality (IoU) would improve if it was completely corrected.
We use this to create annotator corrections on the most important error regions (the exact way depends on the particular experiment, details in Sec.~\ref{sec:results}).

To correct an error, we need a scribble that starts inside the ground-truth region and extends into the error region (Fig.~\ref{fig:main_idea}).
We simulate such scribbles with a three-step process, illustrated in Fig.~\ref{fig:error_region}:
(1) first we randomly sample the first point on the border of the error region that touches the ground-truth region (yellow point in Fig.~\ref{fig:error_region};
(2) then we sample two more points uniformly inside the error region (yellow points in Fig.~\ref{fig:error_region}).
(3) we construct a scribble as a smooth trajectory through these three points (using a bezier curve).
We repeat this process ten times, and keep the longest scribble that is exclusively inside the ground-truth region (while all simulated points are within the ground-truth, the curve could cover parts outside the ground-truth).

%% file: results.tex
\vspace{0cm}

\section{Results}\label{sec:results}

We use Mask-RCNN as basic segmentation framework instead of Fully Convolutional architectures~\cite{chen18pami,long15cvpr,ronneberger15miccai} commonly used in single object segmentation works~\cite{hu19nn,le18eccv,li18cvpr,liew17iccv,mahadevan18bmvc,maninis18cvpr,xu16cvpr}. We first demonstrate in Sec.~\ref{sec:dextr} that this is a valid choice by comparing to DEXTR~\cite{maninis18cvpr} in the non-interactive setting where we generate masks starting from extreme points~\cite{papadopoulos17iccv}.
In Sec~\ref{sec:full_image} we move to the full image segmentation task and demonstrate improvements resulting from sharing extreme points across regions and from our new pixel-wise loss.
Finally, in Sec.~\ref{sec:interactive} we show results on interactive full image segmentation, where we also share scribble corrections across regions, and allow the annotator to freely allocate scribbles to regions while considering the whole image.

\vspace{0cm}
\subsection{Single object segmentation}
\label{sec:dextr}

\mypartop{DEXTR}.
In DEXTR~\cite{maninis18cvpr} they predict object masks from four extreme points~\cite{papadopoulos17iccv}.
DEXTR is based on Deeplab-v2~\cite{chen18pami}, using a ResNet-101~\cite{he16cvpr} backbone architecture and a Pyramid Scene Parsing network~\cite{zhao17cvpr} as prediction head. As input they crop a bounding box out of the RGB image based on the extreme points provided by the annotator. The locations of the extreme points are Gaussian blurred and fed as a heatmap to the network, concatenated to the cropped RGB input.
The DEXTR segmentation model obtained state-of-the-art results on this task~\cite{maninis18cvpr}.

\mypar{Details of our model.}
We compare DEXTR to a single object segmentation variant of our model (\single model). It uses the original Mask-RCNN loss, computed individually per mask, and does not share annotations across regions.
For fair comparison to DEXTR, here we also use a ResNet-101~\cite{he16cvpr} backbone, which due to memory constraints limits the resolution of our RoI features to $14 \times 14$ pixels and our predicted mask to $33\times 33$.
Moreover, we use their released code to generate simulated extreme point annotations. In contrast to subsequent experiments, here we also use the same Gaussian blurred heatmaps to input annotations to our model as used in~\cite{maninis18cvpr}.

\mypar{Dataset.}
We follow the experimental setup of~\cite{maninis18cvpr} on the COCO dataset~\cite{lin14eccv}, which has 80 object classes. Models are trained on the 2014 training set and evaluated on the 2017 validation set (previously referred to as 2014 \texttt{minival}). We measure performance in terms of Intersection-over-Union averaged over all instances.

\begin{table}[t]
\centering
\vspace{-0.2cm}
\begin{tabular}{lc}
\toprule
Method  & IoU\\ \midrule
DEXTR \cite{maninis18cvpr} & 82.1\\
DEXTR (released model) & 81.9\\
Our \single model & 81.6\\
\bottomrule
\end{tabular}
\caption{\small{Performance on COCO (objects only). The accuracy of our \single model is comparable to DEXTR~\cite{maninis18cvpr}.}}
\label{tab:dextr}
\end{table}
\begin{table}[t]
\centering
\vspace{0cm}
\begin{tabular}{@{}lcc@{}}
\toprule
               & X-points not shared & X-points shared \\
\midrule
Mask-wise loss  & 75.8     & 76.0        \\
Pixel-wise loss & 78.4     & 79.1        \\
\bottomrule
\end{tabular}
\caption{\small{Performance on the COCO Panoptic validation set when predicting masks from extreme points (X-points). We vary the loss and whether extreme points are shared across regions. The top-left entry corresponds to our \single model, the bottom-right entry corresponds to our \full model.}}
\label{tab:ablation}
\end{table}

\mypar{Results.}
Tab.~\ref{tab:dextr} reports the original results from DEXTR~\cite{maninis18cvpr}, our reproduction using their publicly released model, and results of our \single model. Their publicly released model and our model deliver very similar results (81.9 and 81.6 IoU). This demonstrate that Mask-RCNN-style models are competitive to commonly used FCN-style models for this task.

\vspace{0cm}
\subsection{Full image segmentation}\label{sec:full_image}

\mypartop{Experimental setup.}
Given extreme points for each object and \stuff region, we now predict a full image segmentation. We demonstrate the benefits of using our pixel-wise loss (Sec.~\ref{sec:loss}) and sharing extreme points across regions (i.e. extreme points for one region are used as negative information for nearby regions, Sec.~\ref{sec:annotation_features}).

\mypar{Details of our model.}
In preliminary experiments we found that RoI features of $14 \times 14$ pixels resolution were limiting accuracy when feeding in annotations into the segmentation head. Therefore we increased both the RoI features and the predicted mask to $41 \times 41$ pixels and switched to ResNet-50~\cite{he16cvpr} due to memory constraints. Importantly, in all experiments from now on our model uses the two-channel annotation maps described in Sec.~\ref{sec:annotation_features}.

\mypar{Dataset.}
We perform our experiments on the COCO panoptic challenge dataset~\cite{caesar18cvpr,kirillov18arxiv,lin14eccv}, which has 80 object classes and 53 stuff classes. Since the final goal is to efficiently annotate data, we train on only 12.5\% of the 2017 training set (15k images). We evaluate on the 2017 validation set and measure IoU averaged over all object and stuff regions in all images.

\mypar{Results.}
As Tab.~\ref{tab:ablation} shows, our \single model yields 75.8 IoU. It uses a mask-wise loss and does not share extreme points across regions.
When only sharing extreme points, we get a small gain of +0.2 IoU.
In contrast, when only switching to our pixel-wise loss, results improve by +2.6 IoU. Sharing extreme points is more beneficial in combination with our new loss, yielding an additional improvement of +0.7 IoU.
Overall this model with both improvements achieves 79.1 IoU, +3.3 higher than the \single model. We call it our \full model.

\subsection{Interactive full image segmentation}
\label{sec:interactive}

We now move to our final system for interactive full image segmentation. We start from the segmentations from extreme points made by our \single and \full models from Sec.~\ref{sec:full_image}.
Then we iterate between:
(A) adding scribble corrections by the annotator, and
(B) updating the machine segmentations accordingly.

\mypar{Dataset and training.}
As before, we experiment on the COCO panoptic challenge dataset and report results on the 2017 validation set. Since during iterations our models input scribble corrections in addition to extreme points, we train two new interactive models: \singlescrib and \fullscrib. These models have the same architecture as their counterparts in Sec.~\ref{sec:full_image} (which input only extreme points), but are trained differently.
To create training data for one of these interactive models, we apply its counterpart to another 12.5\% of the 2017 training set.
We generate simulated corrective scribbles as described in Sec.~\ref{sec:annotator} and train each model on the combined extreme points and scribbles annotations (Sec.~\ref{sec:annotation_features}).
We keep these models fixed throughout all iterations of interactive segmentation.
Note how, in addition to sharing extreme points as in Sec.~\ref{sec:full_image}, the \fullscrib model also shares scribble corrections across regions.

\mypar{Allocation of scribble corrections.}
When using our \singlescrib model, in every iteration we allocate exactly one scribble to each region.
Instead, when using our \fullscrib model we also consider an alternative interesting strategy:
one scribble per region {\em on average}, but the annotator can freely allocate these scribbles to the regions in the image.
This enables the annotator to focus efforts on the biggest errors across the whole image, typically resulting in some regions receiving multiple scribbles and some receiving none.

\begin{figure}
\centering
\vspace{-0.4cm}
\hspace*{-0.5cm}
\vspace{-0.2cm}
\includegraphics[width=0.54\textwidth]{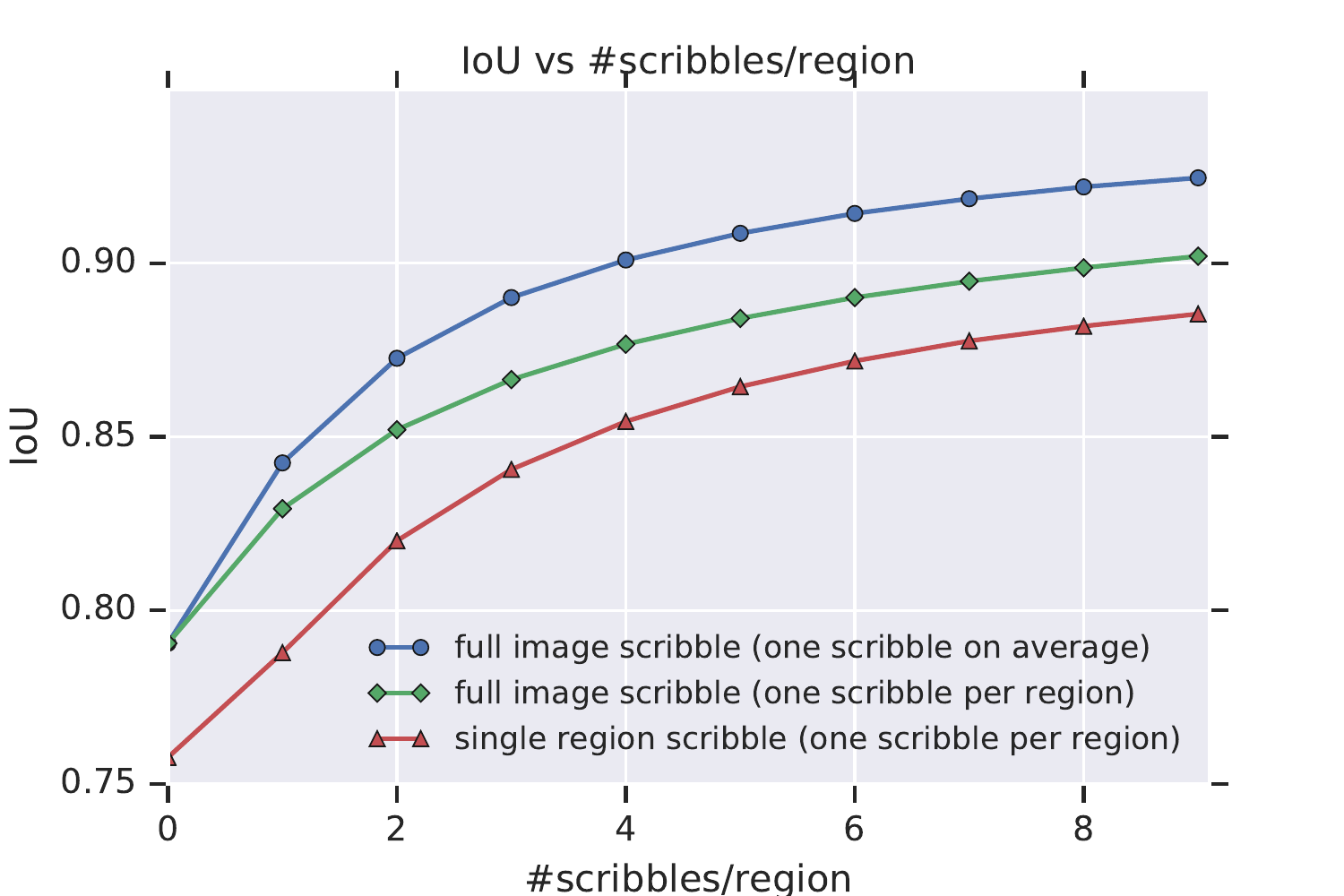}
\caption{\small{Results on the COCO Panoptic validation set for the interactive full image segmentation task. We measure average IoU vs the number of scribbles per region. We compare our \fullscrib model under two scribble allocation strategies to the \singlescrib baseline.}}
\vspace{-.0cm}
\label{fig:main_results}
\end{figure}

\newcommand{\exwidth}{3.00cm}
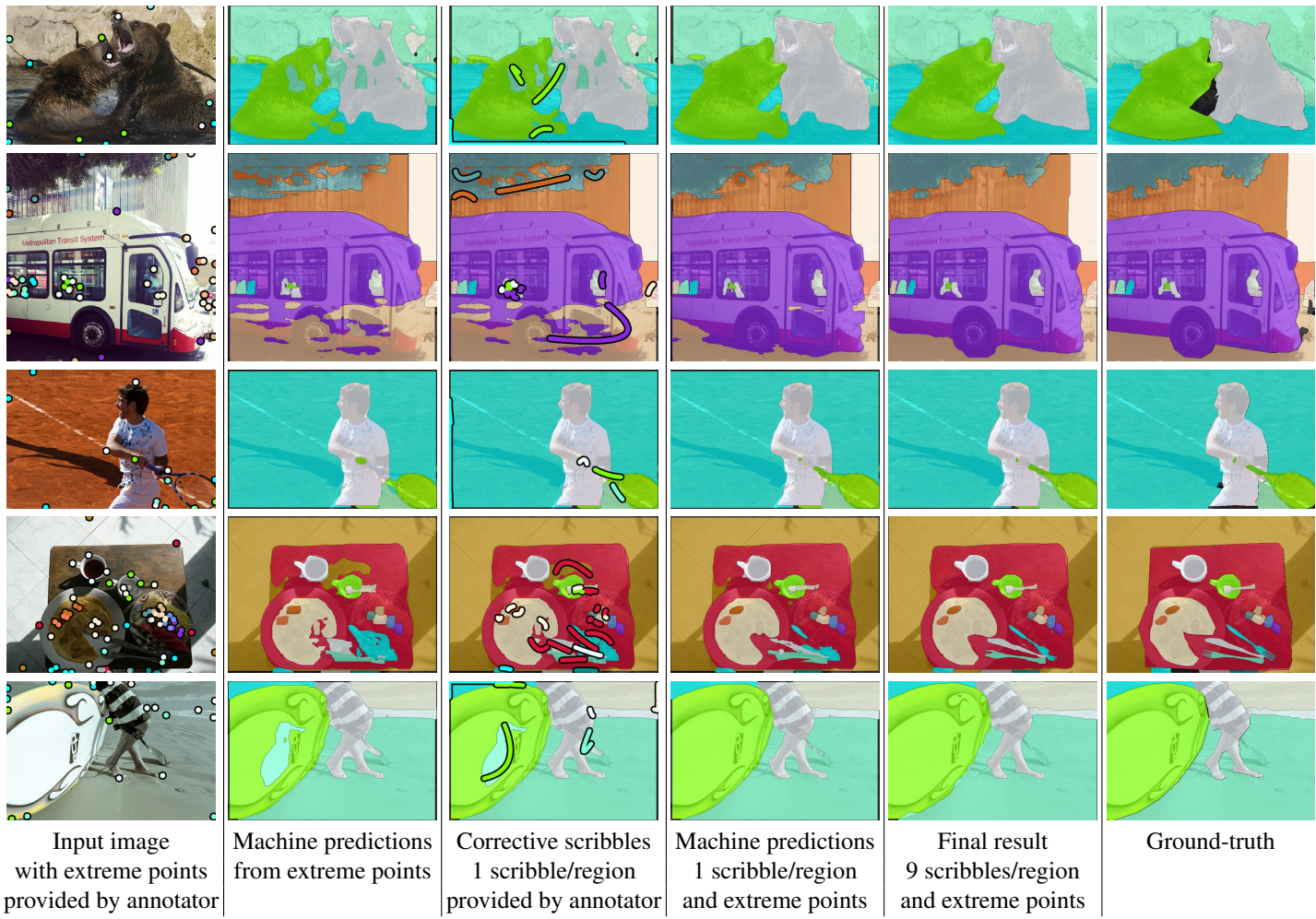
\begin{figure*}
    \setlength{\tabcolsep}{2pt} %
\vspace*{0.2cm}
\hspace*{-0.6cm}
\begin{tabular}{c|c|c|c|c|c}
	\input{examples}
	Input image& 
	Machine predictions&
	Corrective scribbles&
	Machine predictions&
	Final result&
	Ground-truth\\
	with extreme points&
	from extreme points&
	1 scribble/region&
	1 scribble/region&
	9 scribbles/region\\
    provided by annotator & & provided by annotator & and extreme points & and extreme points
\end{tabular}
    \vspace{-0.0cm}
    \caption{\small{We show example results obtained by our system using the $\fullscrib$ model with a free allocation strategy. The first two columns show the input image with extreme points and predictions. Column 3 shows the first annotation step with one scribble correction per region on average, and column 4 shows the updated predictions. The last two columns compare the final result after 9 steps (using 9 scribbles per region on average) with the COCO ground-truth segmentation.}}
   \label{fig:qualitative_examples}
   \vspace{-0.0cm}
\end{figure*}

\vspace{0cm}
\mypar{Results.}
Fig.~\ref{fig:main_results} shows annotation quality (IoU) vs cost (number of scribbles per region).
The two starting points at zero scribbles are the same as the top-left and bottom-right entries of Tab.~\ref{tab:ablation} since they are made using the same non-interactive models (from extreme points only).

We first compare \singlescrib to \fullscrib while using the same allocation strategy: exactly one scribble per region.
Fig.~\ref{fig:main_results} shows that for both models accuracy rapidly improves with more scribble corrections. However, \fullscrib always offers a better trade-off between annotation effort and segmentation quality, e.g. to reach 85\% IoU it takes 4 scribbles per region for the \singlescrib model but only 2 scribbles for our \fullscrib model.
Similarly, to reach 88\% IoU it takes 7 scribbles vs 4 scribbles.
This confirms that the benefits of sharing annotations across regions and of our pixel-wise loss persist also in the interactive setting.

We now compare the two scribble allocation strategies on the \fullscrib model.
As Fig.~\ref{fig:main_results} shows, using the strategy of freely allocating scribbles to regions (one scribble on average) brings further efficiency gains. \fullscrib reaches a very high 90\% IoU at just 4 scribbles per region on average. Reaching this IoU instead requires allocating exactly 8 scribbles per region with the other strategy.
This demonstrate the benefits of focusing annotation effort on the largest errors across the whole image.

Overall, at a budget of four extreme clicks and four scribbles per region, we get a total 5\% IoU gain over \singlescrib (90\% vs 85\%). This gain is brought by the combined effects of our contributions: sharing annotations across regions, the pixel-wise loss which lets regions compete on the common image canvas, and the free scribble allocation strategy.

Fig.~\ref{fig:qualitative_examples} shows various examples for how annotation progresses over iterations. Notice how in the first example, the corrective scribble on the left bear induces a negative scribble for the rock, which in turn improves the segmentation of the right bear. This demonstrates the benefit of sharing scribble annotations and competition between regions.

%% file: examples.tex
	\includegraphics[width=\exwidth]{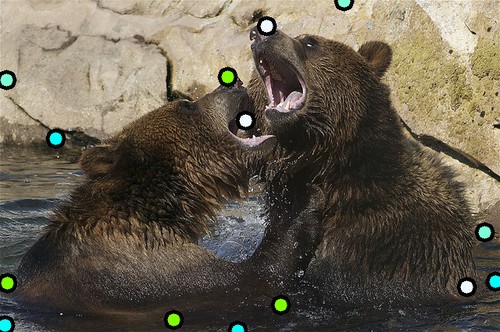}&
	\includegraphics[width=\exwidth]{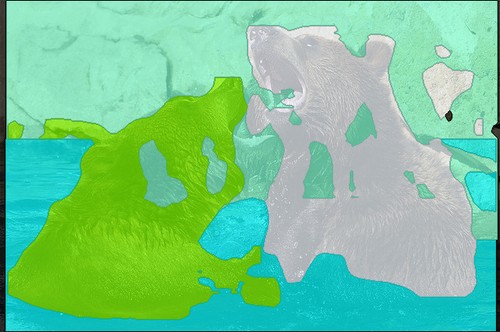}&
	\includegraphics[width=\exwidth]{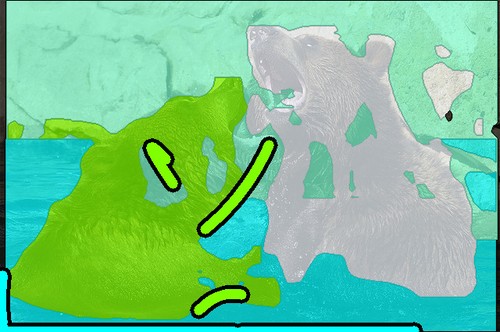}&
	\includegraphics[width=\exwidth]{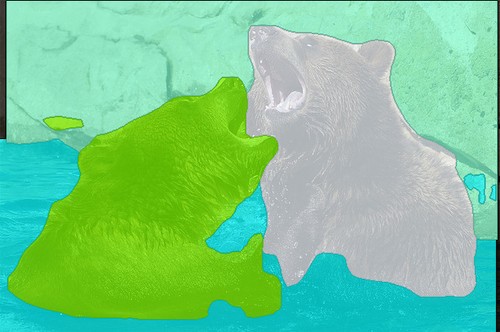}&
	\includegraphics[width=\exwidth]{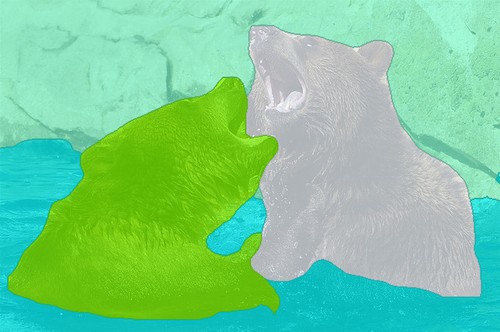}&
	\includegraphics[width=\exwidth]{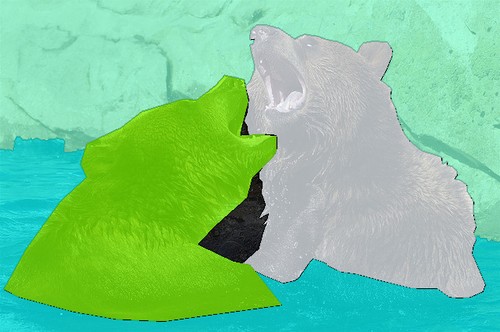}\\
	\includegraphics[width=\exwidth]{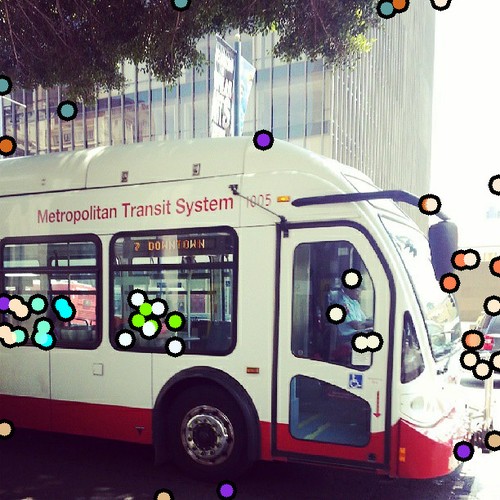}&
	\includegraphics[width=\exwidth]{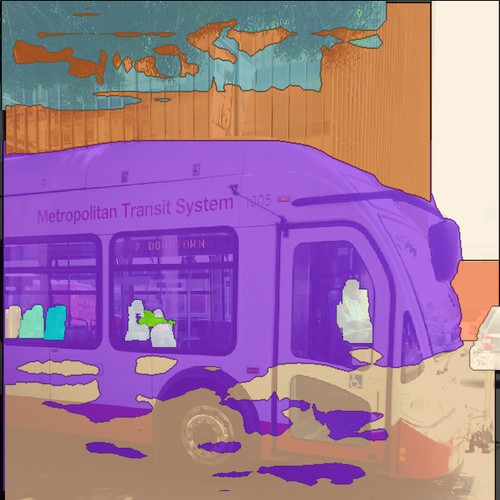}&
	\includegraphics[width=\exwidth]{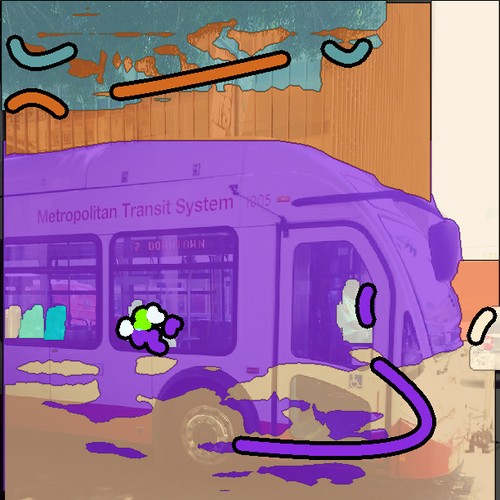}&
	\includegraphics[width=\exwidth]{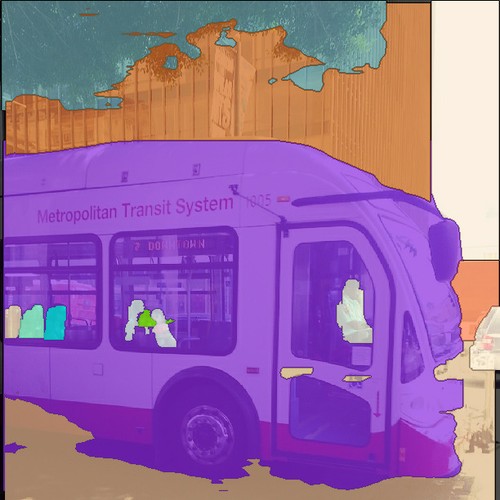}&
	\includegraphics[width=\exwidth]{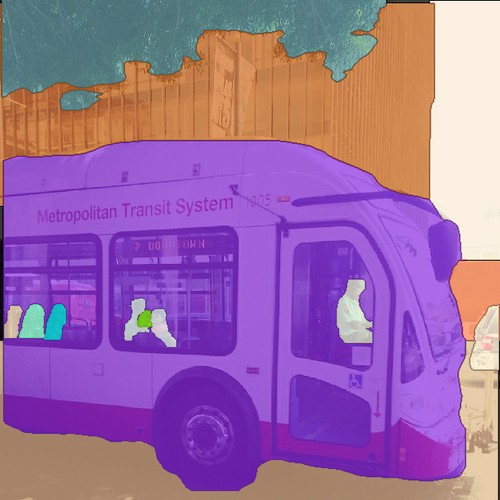}&
	\includegraphics[width=\exwidth]{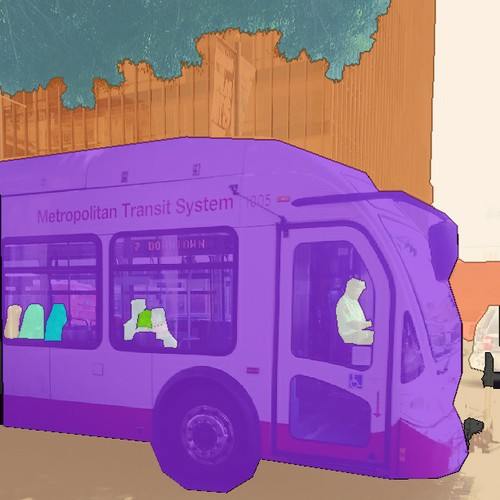}\\
	\includegraphics[width=\exwidth]{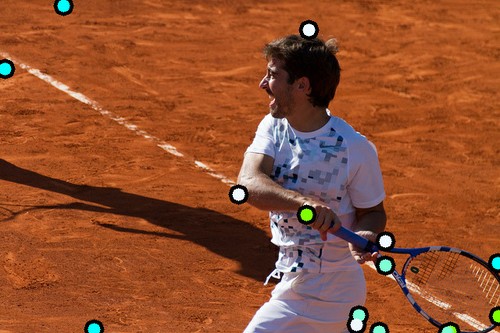}&
	\includegraphics[width=\exwidth]{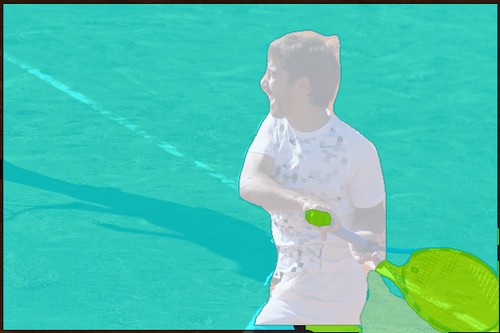}&
	\includegraphics[width=\exwidth]{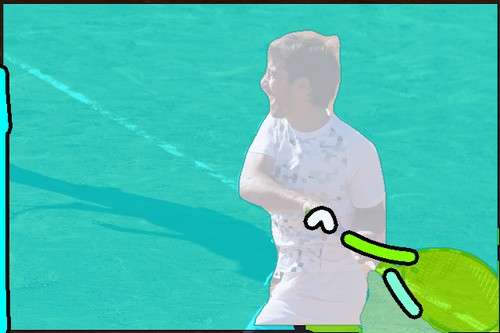}&
	\includegraphics[width=\exwidth]{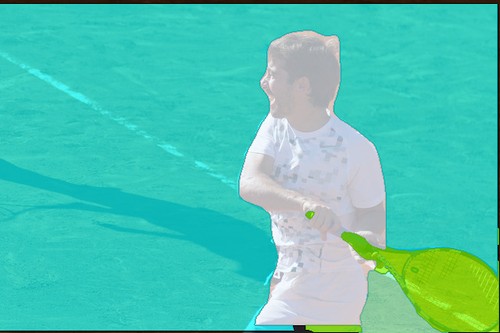}&
	\includegraphics[width=\exwidth]{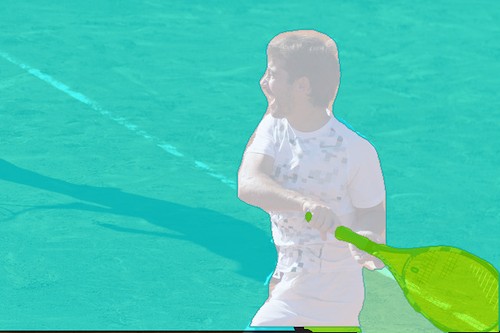}&
	\includegraphics[width=\exwidth]{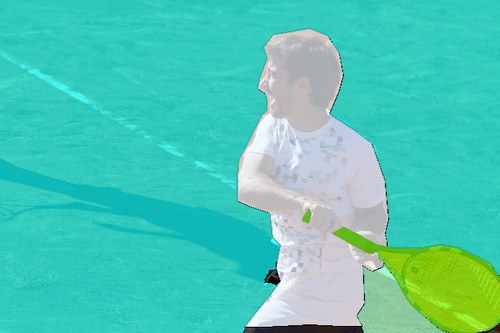}\\
	\includegraphics[width=\exwidth]{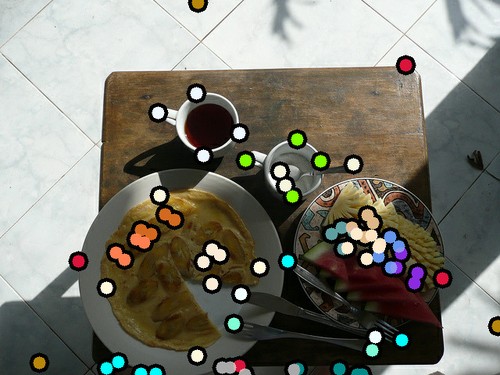}&
	\includegraphics[width=\exwidth]{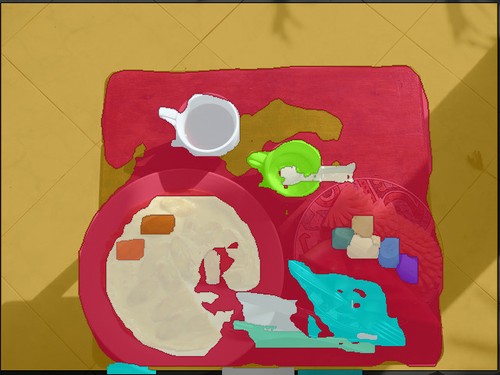}&
	\includegraphics[width=\exwidth]{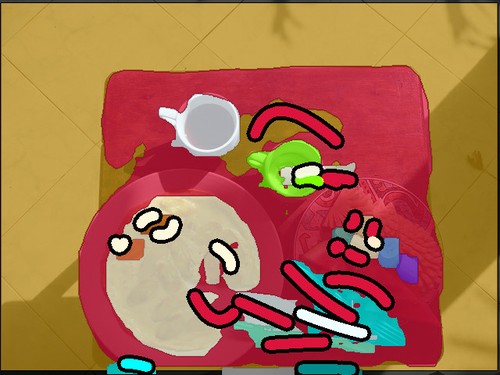}&
	\includegraphics[width=\exwidth]{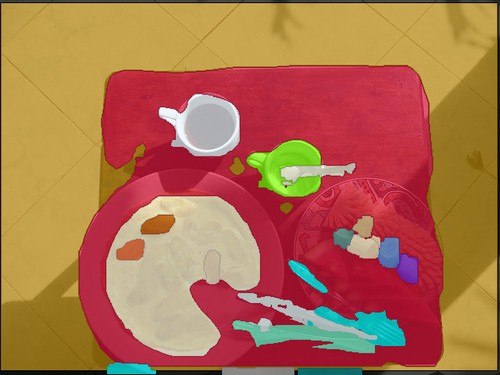}&
	\includegraphics[width=\exwidth]{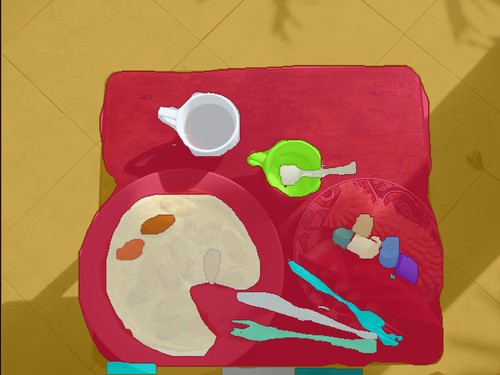}&
	\includegraphics[width=\exwidth]{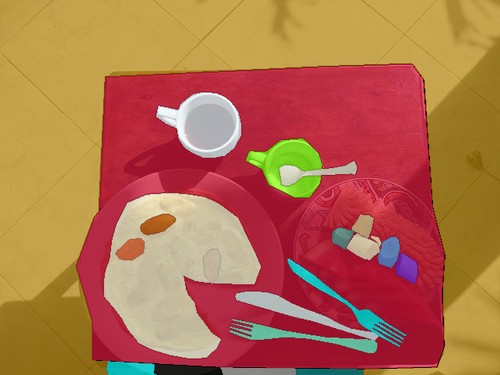}\\
	\includegraphics[width=\exwidth]{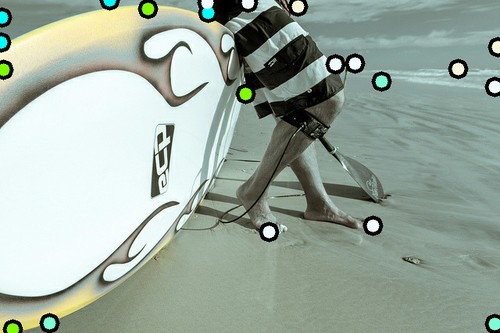}&
	\includegraphics[width=\exwidth]{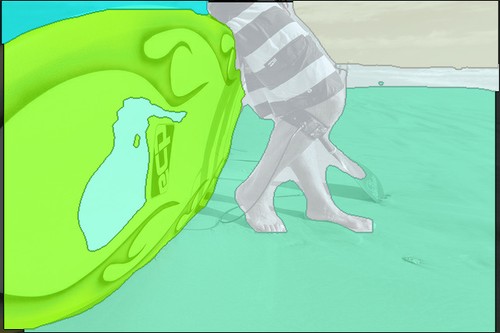}&
	\includegraphics[width=\exwidth]{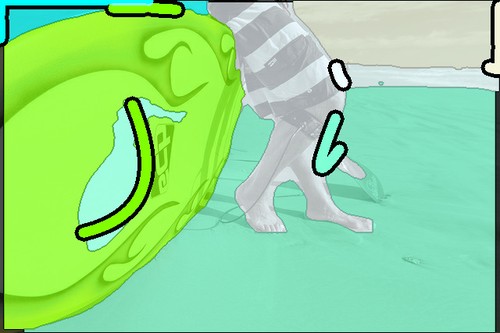}&
	\includegraphics[width=\exwidth]{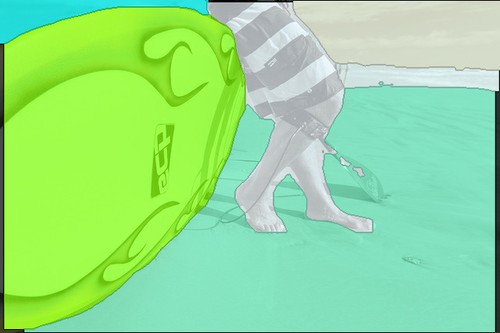}&
	\includegraphics[width=\exwidth]{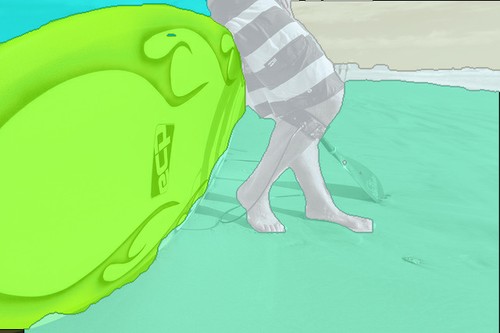}&
	\includegraphics[width=\exwidth]{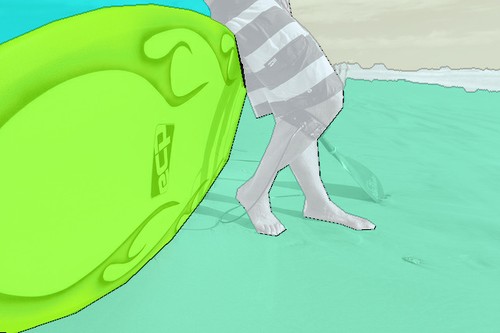}\\

%% file: discussion.tex
\section{Discussion}\label{sec:discussion}

\mypartop{Mask-RCNN vs FCNs.}
Our work builds on Mask-RCNN~\cite{he17iccv} rather than FCN-based models~\cite{chen18pami,long15cvpr,ronneberger15miccai} because it is faster and requires less memory. To see this, we can reinterpret Fig.~\ref{fig:architecture} as an FCN-based model: ignore the backbone network, replace the backbone features $\features$ by the RGB image, and make the segmentation head a full FCN.

At inference time, we need to do a forward pass through the segmentation head for every region for every correction. When using Mask-RCNN, the heavy ResNet~\cite{he16cvpr} backbone network is applied only once for the whole image, and then only a small 4-layer segmentation head is applied to each region. For the FCN-style alternative instead, nothing can be precomputed and the segmentation head itself is the heavy ResNet. Hence our framework is much faster during interactive annotation.

During training, typically all intermediate network activations are stored in memory. Crucially, for each region distinct activations are generated in the segmentation head. For FCN-style models this is a heavy ResNet and requires lots of memory. This is why DEXTR~\cite{maninis18cvpr} reports a maximal batch size of 5 regions.
Therefore, it would be difficult to train with our pixel-wise loss in an FCN-style model, as that requires processing all regions in each image simultaneously (15 regions per image on average).

In fact our Mask-RCNN based architecture (Fig.~\ref{fig:architecture}) and its reinterpretation as an FCN-based model span a continuum. Its design space can be explored by varying the size of the backbone and the segmentation head, as well as their input and output resolution. We leave such exploration of the trade-off between memory requirements, inference speed, and model accuracy for future work.

\mypartop{Scribble and point simulations.}
Like other interactive segmentation works~\cite{acuna18cvpr,castrejon17cvpr,hu19nn,le18eccv,li18cvpr,liew17iccv,mahadevan18bmvc,maninis18cvpr,xu16cvpr}, we simulate annotations.
It remains to be studied how to best select the simulation parameters so that the models generalize well to real human annotators.
The optimal parameters will likely depend on various factors, such as the desired annotation quality and the accuracy of the provided corrections.

%% file: conclusion.tex
\section{Conclusions}

We proposed an interactive annotation framework which operates on the whole image to produce segmentations for all object and \stuff regions. Our key contributions derive from considering the full image at once: sharing annotations across regions, focusing annotator effort on the biggest errors across the whole image, and a pixel-wise loss for Mask-RCNN that lets regions compete on the common image canvas.
We have shown through experiments on the COCO panoptic challenge dataset~\cite{caesar18cvpr,kirillov18arxiv,lin14eccv} that all the elements we propose improve the trade-off between annotation cost and quality, leading to a very high IoU of 90\% using just four extreme points and four corrective scribbles per region (compared to 85\% for the baseline).